\documentclass[letterpaper, 10 pt, conference]{ieeeconf}  

\IEEEoverridecommandlockouts                              

\usepackage{amssymb}

\usepackage{amsthm}

\usepackage{amsmath}
\usepackage{bm}
\usepackage{graphicx}

\usepackage{subcaption}

\usepackage{hyperref}
\usepackage{algorithm}
\usepackage{algorithmic}

\usepackage[rightcaption]{sidecap}
\usepackage{multirow} 
\usepackage{tcolorbox}

\let\labelindent\relax
\usepackage{enumitem}
\setlist[itemize]{leftmargin=*, noitemsep}
\setlist[enumerate]{leftmargin=*, noitemsep}
\usepackage{booktabs}
\usepackage{listings}

\usepackage{xcolor}
\usepackage{adjustbox} 
\usepackage{colortbl} 
\definecolor{lightgray}{gray}{0.9} 
\usepackage{mathtools} 
\usepackage{makecell}

\usepackage[T1]{fontenc}
\usepackage{lmodern}
\newcommand{\meanstd}[2]{#1\textcolor{gray}{\textsubscript{$\pm$#2}}}

\title{\LARGE \bf
Risk-Aware Human-in-the-Loop Framework with Adaptive Intrusion Response for Autonomous Vehicles}

\author{%
\begin{tabular}{c}
\textbf{Dawood Wasif$^{1}$, Terrence J. Moore$^{2}$, Seunghyun Yoon$^{3}$, Hyuk Lim$^{3}$,}\\
\textbf{Dan Dongseong Kim$^{4}$, Frederica F. Nelson$^{2}$, Jin-Hee Cho$^{1}$}\\[2mm]
{\small $^{1}$Virginia Tech, USA \quad $^{2}$U.S. Army Research Laboratory, USA}\\
{\small $^{3}$KENTECH, Republic of Korea \quad $^{4}$The University of Queensland, Australia}
\end{tabular}
}

\begin{document}

\maketitle
\thispagestyle{empty}
\pagestyle{empty}

\begin{abstract}
Autonomous vehicles must remain safe and effective when encountering rare long-tailed scenarios or cyber–physical intrusions during driving. We present \underline{R}\underline{A}\underline{I}\underline{L}, a risk-aware human-in-the-loop framework that turns heterogeneous runtime signals into calibrated control adaptations and focused learning. RAIL fuses three cues (curvature actuation integrity, time-to-collision proximity, and observation-shift consistency) into an Intrusion Risk Score (IRS) via a weighted Noisy-OR. When IRS exceeds a threshold, actions are blended with a cue-specific shield using a learned authority, while human override remains available; when risk is low, the nominal policy executes. A \emph{contextual bandit} arbitrates among shields based on the cue vector, improving mitigation choices online. RAIL couples Soft Actor–Critic (SAC) with risk-prioritized replay and dual rewards so that takeovers and near misses steer learning while nominal behavior remains covered. On MetaDrive, RAIL achieves a \emph{Test Return (TR)} of 360.65, a \emph{Test Success Rate (TSR)} of 0.85, a \emph{Test Safety Violation (TSV)} of 0.75, and a \emph{Disturbance Rate (DR)} of 0.0027, while logging only 29.07 training safety violations—outperforming RL, safe RL, offline/imitation learning, and prior HITL baselines. Under Controller Area Network (CAN) injection and LiDAR spoofing attacks, it improves \emph{Success Rate (SR)} to 0.68 and 0.80, lowers the \emph{Disengagement Rate under Attack (DRA)} to 0.37 and 0.03, and reduces the \emph{Attack Success Rate (ASR)} to 0.34 and 0.11. In CARLA, RAIL attains a TR of 1609.70 and TSR of 0.41 with only 8K steps.
\end{abstract}

\section{INTRODUCTION}

Autonomous vehicles (AVs) are moving from controlled pilots to open-world deployment, promising safer and more efficient transportation by offloading perception, planning, and control from humans to machine intelligence. Progress in large-scale simulation, high-capacity function approximation, and model-based and model-free control has produced competent lane keeping, merging, and urban navigation, often with performance that approaches human driving in benign conditions \cite{parekh2022review}. As the autonomy stack matures, the community has moved from “can it drive” to “when and why does it fail,” with emphasis on sample efficiency, transfer to unseen maps, and transparent interaction with human operators. This progress raises a central requirement for deployment: AVs must reason about safety and uncertainty while coordinating seamlessly with human oversight.

The general problem that emerges in safety-critical autonomy is that learned policies are exposed to adversarial perturbations or rare long-tail scenarios that are not fully captured during development. Pure reinforcement learning (RL) can optimize returns in a simulator, yet it may violate safety constraints during exploration. Safe RL methods introduce constraints but often depend on hand-tuned costs and may underperform when the cost signals are sparse or delayed \cite{achiam2017constrained,stooke2020responsive}. Imitation learning (IL) jumpstarts behavior but can overfit to the logged distribution and degrade when the environment deviates from training data \cite{bain1995framework,ho2016generative}. Offline RL reduces the risk of data collection but inherits dataset bias and imperfect coverage \cite{kumar2020conservative}. Human-in-the-loop (HITL) learning injects expert knowledge through demonstrations, interventions, and shared control \cite{huang2024human}, yet HITL is largely reactive and fragmented, treating human input as static supervision that does not scale well across diverse environments and attack surfaces. Moreover, in most HITL approaches, when a human takes over, most systems treat the correction as an ad hoc event rather than a structured signal to improve the policy \cite{kelly2019hg, saunders2017trial, li2022efficient}.

Significant efforts have attempted to close these gaps. Deep RL baselines such as Soft Actor–Critic (SAC) and Proximal Policy Optimization (PPO) provide data-efficient policy learning for continuous control \cite{haarnoja2018soft,schulman2017proximal}. Safe RL extends these algorithms with constraint satisfaction and tail-risk objectives \cite{tamar2015optimizing,greenberg2022efficient,queeney2023risk,yang2024risk}. HITL frameworks enable intervention-aware training, copilot optimization, and curriculum shaping \cite{li2022efficient,huang2024human}. Intrusion detection for vehicular networks and sensors advances the state of the art in anomaly detection and runtime monitoring \cite{nagarajan2023machine,cho2016fingerprinting,ren2020adversarial,aloraini2024adversarial,sultana2024detects}. Despite substantial progress, practical systems still exhibit important gaps: detection is often decoupled from control, so elevated concern does not translate into calibrated, graded actions at the control rate; human oversight sits outside the primary loop, leaving takeovers and near misses underused as structured supervision; and mitigation logic is largely static, relying on fixed responses rather than adapting with experience in the operator’s domain.

This paper introduces \underline{R}isk-\underline{A}ware \underline{H}uman-\underline{i}n-the-\underline{L}oop with Adaptive Intrusion Response (RAIL), a unified framework for autonomous vehicles that integrates intrusion response and learning into runtime decision making. RAIL couples the control policy with a probabilistic measure of operational concern and converts that concern into interpretable, cue-specific control adjustments blended with the nominal action. A lightweight contextual arbitration mechanism selects which adjustment to apply when concern is high and tunes its strength online based on a success signal correlated with reduced human burden. The learning loop treats human takeovers and safety events as explicit penalties while adding a shaping term that promotes low-concern behavior even before failures occur. A prioritized replay buffer concentrates gradient steps on the most safety-critical slices of experience. Together, these components create a closed loop where detection informs control immediately, choices remain interpretable, and experience continuously improves both the policy and the response mechanism.

\begin{figure}[t]
  \centering
  \includegraphics[width=\columnwidth]{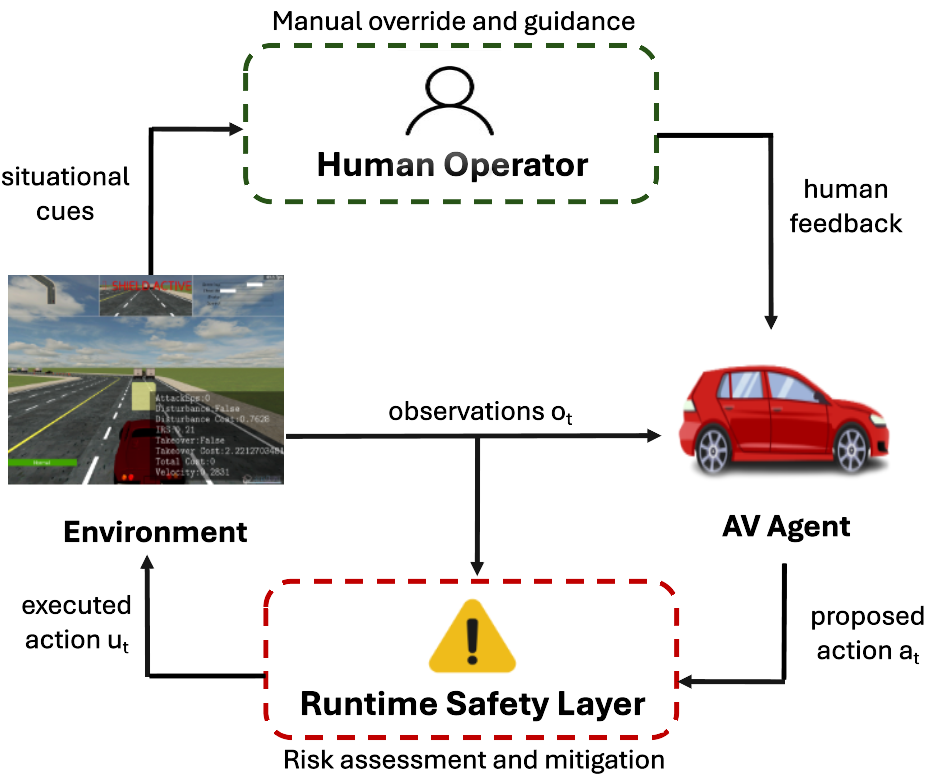} 
 \caption{Paradigm of Risk-Aware HITL Learning}
\label{fig:paradigm}
\end{figure}
Fig.~\ref{fig:paradigm} illustrates the high-level paradigm for risk-aware human-in-the-loop learning. The AV agent proposes an action that is first assessed by a runtime safety layer, which mitigates risk before execution. The environment then returns observations, and a human operator can provide situational cues, manual overrides, and feedback. This closed loop ensures that elevated risk is translated into calibrated responses rather than unsafe actuation.

We summarize our \textbf{key contributions} as follows:
\begin{itemize}
    \item We present the \underline{R}isk-\underline{A}ware \underline{H}uman-\underline{i}n-the-\underline{L}oop (RAIL) framework, a unified runtime architecture that links detection, graded response, and human escalation in a single decision loop operating at control rate and logging outcomes for learning.
    \item We introduce the Intrusion Risk Score (IRS), a probabilistic measure that fuses three runtime cues—\textit{curvature uncertainty} (plan–execute mismatch), \textit{time-to-collision} proximity, and \textit{LiDAR observation-shift} (OOD) detection—into a single value, mapping it to cue-specific, transparent control adjustments that operators can audit, with a contextual selection mechanism that tunes adjustment strength based on current signals and delayed success indicators.
    \item We advance safety-aware learning by coupling explicit penalties for violations and takeovers with an implicit shaping term that promotes low-concern behavior, and by proposing a replay scheme that prioritizes high-concern and takeover transitions while preserving diversity within an off-policy actor-critic backbone.
    \item We evaluate RAIL extensively in MetaDrive and CARLA, demonstrating improved safety, robustness, and operator efficiency over RL, safe RL, imitation, offline RL, and prior human-in-the-loop baselines, under both nominal conditions and representative cyber–physical disturbances.
\end{itemize}


\section{Related Work}

\subsection{Human-in-the-Loop (HITL) Learning Methods}
HITL learning integrates human demonstrations, oversight, and intervention with data-driven control to stabilize training and improve safety in sequential decision-making. In autonomous driving, HITL sits alongside modern deep reinforcement learning approaches such as SAC and PPO that provide strong baselines for continuous control \cite{haarnoja2018soft,schulman2017proximal}, and it complements safe RL methods that enforce constraints during optimization, including constrained policy optimization and PID Lagrangian techniques \cite{achiam2017constrained,stooke2020responsive}. Supervised and adversarial imitation methods supply initial competencies from logs, from classical behavioral cloning to distribution-matching via generative adversarial imitation learning \cite{bain1995framework,ho2016generative}, while interactive variants collect corrective labels online through expert-in-the-loop aggregation and teleoperation \cite{kelly2019hg,mandlekar2020human}. When direct online exploration is costly or risky, offline RL methods like conservative Q-learning exploit static datasets \cite{kumar2020conservative}, and recent work demonstrates learning with minimal human effort in real systems as well as shared-autonomy copilot optimization in driving \cite{huang2024human,li2022efficient, ha2020learning}. Despite this progress, many HITL pipelines remain fragmented across learning, safety, and oversight channels, with limited mechanisms to integrate diverse signals and act consistently under uncertainty.

\subsection{Risk-Aware Decision Making}
Risk-sensitive reinforcement learning methods model the tails of return distributions using coherent risk measures  and distributional critics to attenuate catastrophic outcomes \cite{tamar2015optimizing, greenberg2022efficient}. Surveys emphasize that risk-aware policy optimization is critical for safety-critical robotics and that combining model uncertainty with distributional value estimates improves reliability \cite{queeney2023risk}. Algorithmic advances include proximal policy gradients with safety/risk constraints and hybrid critics mixing mean and tail objectives \cite{yang2024risk}. Complementary lines impose action filters via runtime shielding or barrier certificates to guarantee constraint satisfaction even under model mismatch \cite{alshiekh2018safe, achiam2017constrained}. Nevertheless, many approaches optimize abstract risk on task rewards, leaving open how to \emph{operationalize} risk from multi-source cues \cite{johansen2014foundations, pendleton2016survey} (e.g., perception anomalies, V2X disagreements) and how to translate elevated risk into interpretable, graded control responses in real time. 

\subsection{Intrusion Detection and Response for AVs}
Early automotive security studies demonstrated the feasibility of remotely compromising ECUs and in-vehicle networks, motivating intrusion detection on CAN via specification-based rules and learning-based anomaly models \cite{nagarajan2023machine, cho2016fingerprinting, tamar2015policy, kastner2023distributional, choudhary2018intrusion}. AV-oriented defenses further introduced runtime monitors that cross-check dynamics, actuator consistency, and state estimation to flag control-relevant anomalies during driving \cite{sultana2024detects}. Concurrently, sensor-level attacks (e.g., LiDAR spoofing, camera adversarial patches) have been demonstrated in the lab and field \cite{aloraini2024adversarial, ren2020adversarial, aloraini2024adversarial}. Recent intrusion \emph{response} proposals advocate real-time containment, graceful degradation, and cross-layer corroboration under attack \cite{hamad2024react, abdo2024avmon}. Despite progress, most work isolates detection from control, offers limited, pre-defined mitigation actions, or lacks human-aware escalation strategies; unified, interpretable response policies that fuse cyber and physical risk cues and adapt online to operator feedback remain underexplored.

\begin{figure*}[t]
\centering
\includegraphics[width=\textwidth]{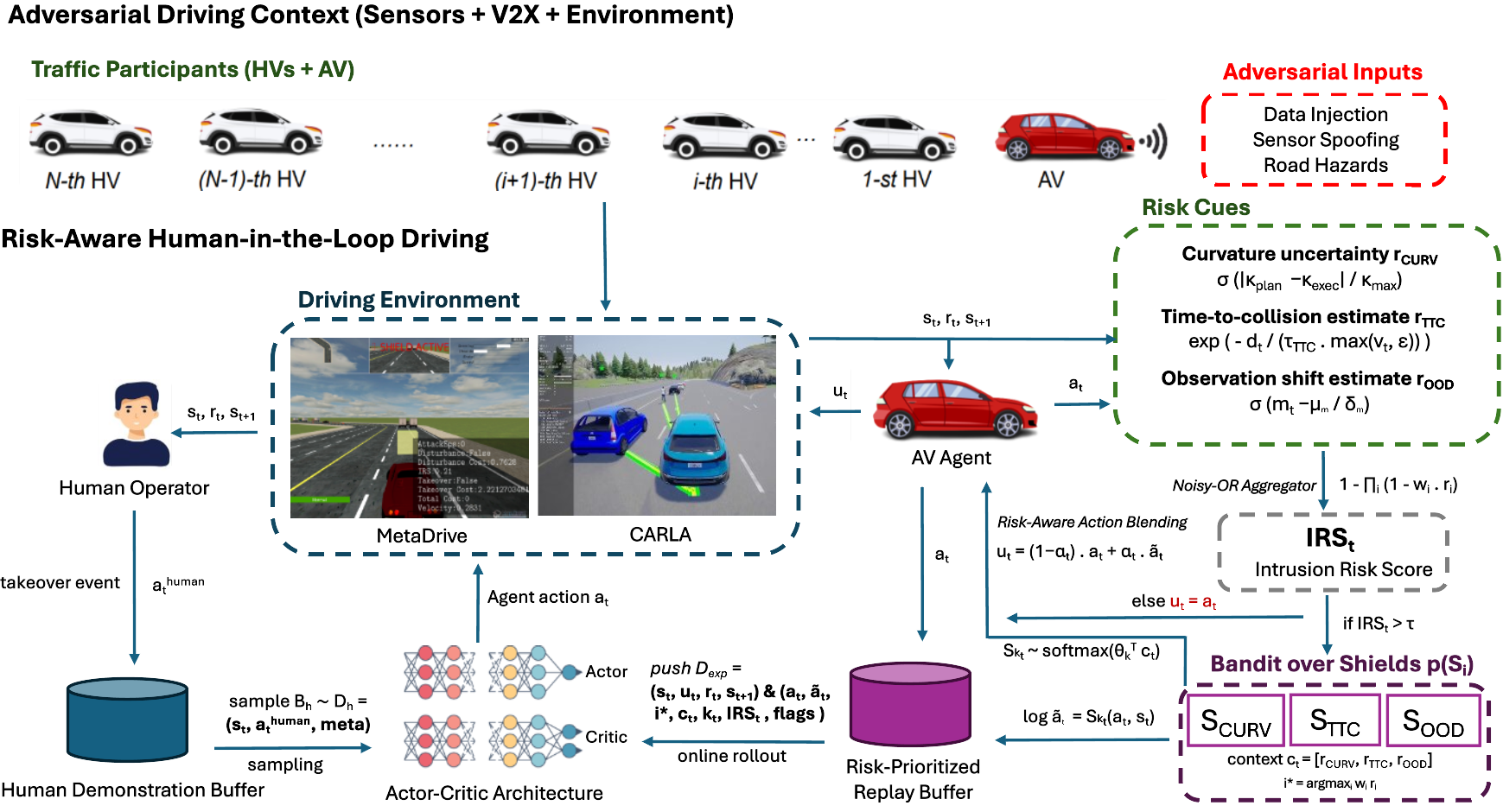}

\caption{RAIL System Overview}
\label{fig:RAIL-arch}
\end{figure*}
\section{Proposed Approach: RAIL}
We introduce \textbf{RAIL} (Risk-Aware Human-in-the-Loop with Adaptive Intrusion Response), which turns runtime risk into graded, interpretable control. The policy proposes an action; a safety layer modulates it, and a human can always overrule.

\subsection{System Overview}
We model the ego vehicle at time $t$ with state $s_t$ (ego kinematics, local map features, perception) and a continuous action $a_t$ (steering, longitudinal) sampled from a stochastic policy $\pi_\theta(a_t\mid s_t)$. The environment evolves under the applied control $u_t$ via a transition $p(s_{t+1}\mid s_t,u_t,\xi_t)$ with disturbances $\xi_t$. RAIL augments this loop with an Intrusion Risk Score $\mathrm{IRS}_t\in[0,1]$ computed from heterogeneous runtime cues. When risk is low, the nominal action executes ($u_t=a_t$). As risk rises, a cue-specific shield produces a safeguarded proposal $\tilde a_t$ and the executor blends it with the policy action; a human override remains available. A lightweight contextual bandit selects the shield and adapts its gain from the current cue vector and delayed success signals. Learning is off-policy with SAC: the replay buffer stores the core transition $(s_t,u_t,r_t,s_{t+1})$, with annotations $(a_t,\tilde a_t,\mathrm{IRS}_t,c_t,i^\star,k_t,\text{flags})$ and prioritizes high-risk or human-corrected transitions so updates focus on safety-critical slices. This architecture couples risk perception, interpretable mitigation, and human supervision within a single closed-loop controller.

\subsection{Intrusion Risk Score (IRS)}

The IRS serves as a probabilistic aggregator of multiple cyber-physical risk cues, transforming heterogeneous measurements into a single interpretable value. Formally, at each time step $t$, we define
\begin{equation}
\label{eq1}
    \text{IRS}_t = 1 - \prod_{i=1}^M \big(1 - w_i r_i(s_t,a_t)\big),
\end{equation}
where $M$ is the number of monitored cues, $r_i(s_t,a_t) \in [0,1]$ denotes the normalized instantaneous risk contribution of cue $i$, and $w_i$ are learned importance weights constrained by $\sum_{i=1}^M w_i = 1$. For interpretability, we record the dominant cue
$i^\star=\arg\max_i w_i r_i(s_t,a_t)$, which indicates the cue contributing most to $\mathrm{IRS}_t$ at time $t$. The multiplicative Noisy-OR form models the probability that at least one cue indicates danger, ensuring that a single high-confidence anomaly (e.g., sudden TTC collapse) is sufficient to elevate $\text{IRS}_t$. Unlike additive scoring, this structure avoids risk underestimation by accounting for inter-cue redundancy and non-linear compounding, effectively capturing how multiple weak but consistent signals may collectively escalate overall risk. Thus, the IRS functions as a latent safety index driving both shield activation and human handover logic. The IRS integrates three ($M=3$) primary cues as follows.

\paragraph{\bf Curvature uncertainty cue}
This cue measures control integrity by comparing the planner’s intended curvature with the vehicle’s executed curvature. The mismatch is converted into a unitless risk in $[0,1]$, with small tracking errors near zero and sustained bias from faults or injected steering near one.

\begin{equation}
r_{\mathrm{CURV}}(s_t,a_t)=\sigma\!\Big(\tfrac{|\kappa_{\mathrm{plan}}(s_t)-\kappa_{\mathrm{exec}}(s_t,a_t)|}{\kappa_{\max}}\Big).
\end{equation}
Here $s_t$ is the state, $a_t=[a_t^{\mathrm{steer}},a_t^{\mathrm{acc}}]$, $\kappa_{\mathrm{plan}}$ is the planned path curvature from route geometry, $\kappa_{\mathrm{exec}}$ is the realized curvature from yaw rate over speed with a small numerical floor, $\kappa_{\max}$ normalizes by feasible limits, and $\sigma$ is a logistic squashing. Risk grows as the executed path diverges from the intended trajectory.

\paragraph{\bf Time-to-collision cue}
This cue serves as a physical proximity check, measuring the imminence of hazard by converting available reaction time into risk. Fabricated obstacles or weakened braking shorten the window and increase risk.
\begin{equation}
r_{\mathrm{TTC}}(s_t,a_t)=\exp\!\Big(-\tfrac{d_t}{\tau_{\mathrm{TTC}}\cdot \max(v_t,\epsilon)}\Big),
\end{equation}
where $d_t$ is the path-aligned distance from the ego bumper to the most threatening object, $v_t$ is the closing speed along the path, $\epsilon>0$ avoids division by zero when the gap is opening, and $\tau_{\mathrm{TTC}}$ sets the time scale. The score is near one when contact is imminent and decays smoothly as the available time increases.

\paragraph{\bf Observation shift cue}
This cue detects sensor-level distribution shift in LiDAR returns and is designed to flag spoofing, masking, or blinding that alters the beam pattern beyond nominal variability.
\begin{equation}
r_{\mathrm{OOD}}(s_t) = 
\sigma\!\left(\frac{\sqrt{(z_t-\mu)^\top(\Sigma+\epsilon I)^{-1}(z_t-\mu)}-\mu_m}{\delta_m}\right),
\end{equation}
where $z_t\!\in\!\mathbb{R}^{72}$ is the LiDAR range vector, $(\mu,\Sigma)$ are the clean-data mean and covariance with Tikhonov regularization $\epsilon I$, and $(\mu_m,\delta_m)$ center and scale the Mahalanobis distance–based score. The risk increases as the current observation becomes less likely under the nominal sensor model.

\subsection{Risk-Aware Action Blending}
RAIL turns the policy’s proposal into executed control by blending it with a cue–specific safety transform only when runtime risk is elevated. Let $a_t$ be the nominal action and $\tilde a_t=\mathcal{S}_{k_t}(a_t,s_t)$ the output of shield $k_t$ selected online (e.g., steering guard, brake bias, speed cap), where $s_t$ is the current state. The executor interpolates with authority $\alpha_t\!\in\![0,1]$ so small risk leaves the policy untouched, while high risk smoothly hands authority to the shield. An exponential moving average smooths the IRS before computing $\alpha_t$, and a rate limiter bounds changes across cycles.

\begin{equation}
u_t=(1-\alpha_t)\,a_t+\alpha_t\,\tilde a_t .
\end{equation}

The shield selection process is cast as a \emph{contextual bandit} over a small library of interpretable transforms. The context is the current cue vector $c_t=[r_{\text{CURV}},r_{\text{TTC}},r_{\text{OOD}}]$. Each arm $k$ carries a linear score $z_{k,t}=\theta_k^\top c_t$ that encodes where that shield is most effective. We sample the arm with a softmax to retain exploration near decision boundaries and to prevent chattering; temperature is annealed over training for convergence.

\begin{equation}
p_t(k)=\frac{\exp(\theta_k^\top c_t)}{\sum_j \exp(\theta_j^\top c_t)} .
\end{equation}

After executing $u_t$, a delayed scalar feedback marks success ($1$ if no takeover within a horizon $\Delta$) and subtracts a small effort penalty $\propto\!\|\tilde a_t-a_t\|$. A lightweight online update nudges $\theta_{k_t}$ in the direction of $c_t$ proportional to this feedback, leaving other arms unchanged; this keeps credit assignment local and preserves interpretability. The design yields (i) \emph{locality}—only the risky channel is modified, (ii) \emph{monotonicity}—shield authority grows with risk, and (iii) \emph{auditability}—the active shield, $\alpha_t$, and scores $\{z_{k,t}\}$ are logged at every step.

\begin{figure*}[t]
\centering
\begin{subfigure}[t]{0.23\textwidth}
  \centering
  \includegraphics[width=\linewidth]{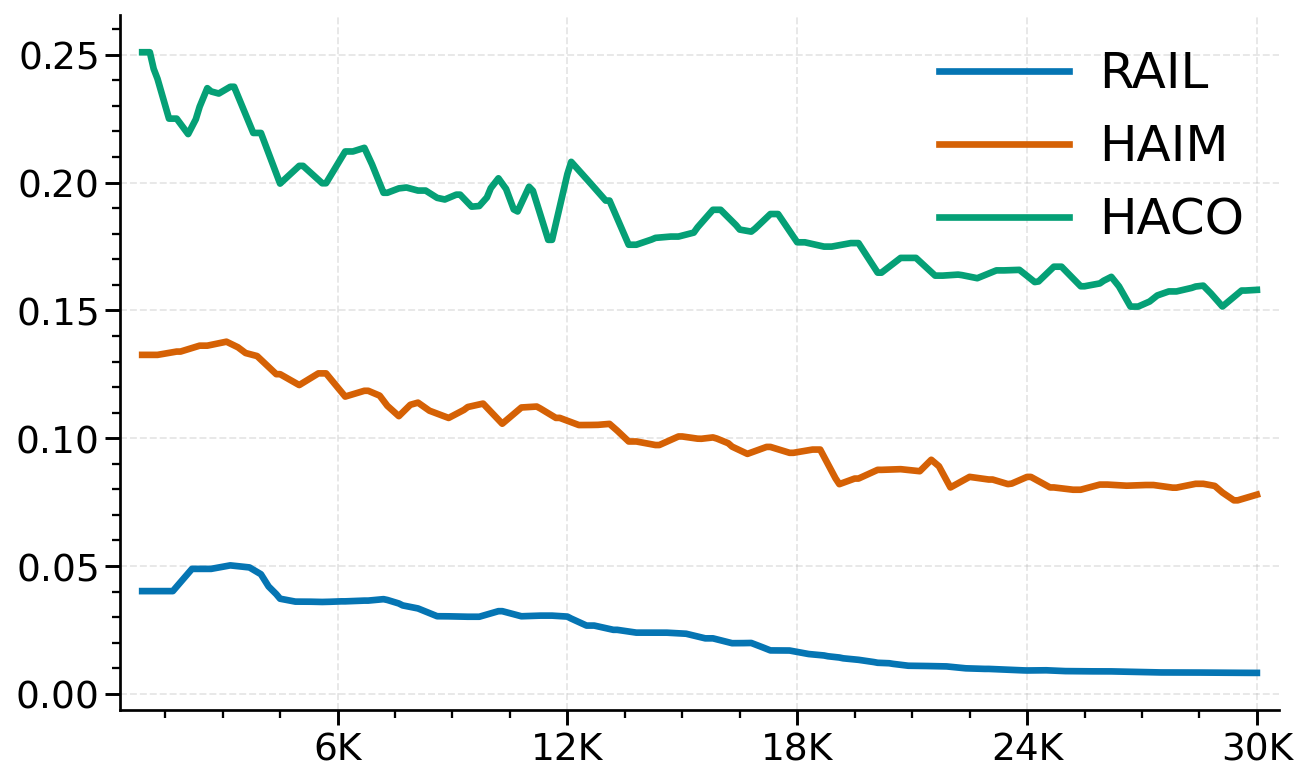}
  \caption{Disturbance rate}
  \label{fig:dist_rate}
\end{subfigure}
\begin{subfigure}[t]{0.23\textwidth}
  \centering
  \includegraphics[width=\linewidth]{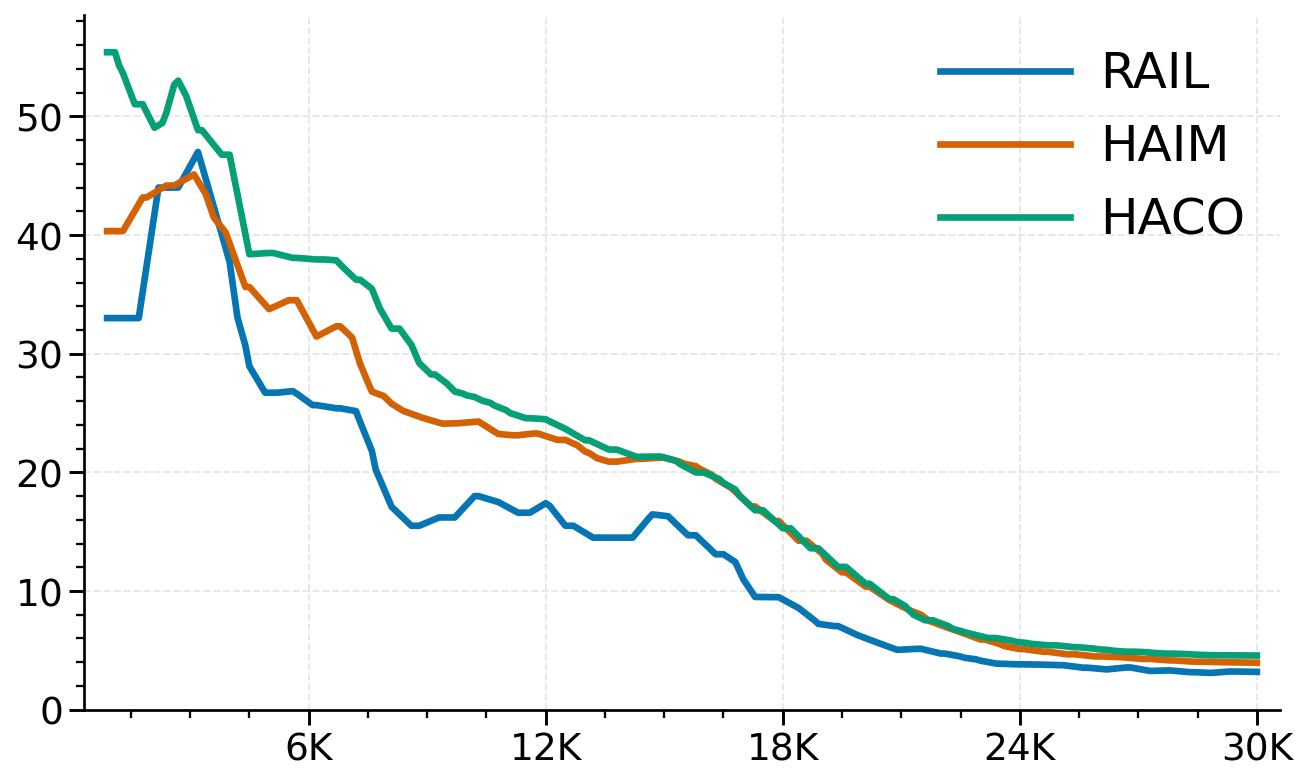}
  \caption{Disturbance count}
  \label{fig:dist_count}
\end{subfigure}
\begin{subfigure}[t]{0.23\textwidth}
  \centering
  \includegraphics[width=\linewidth]{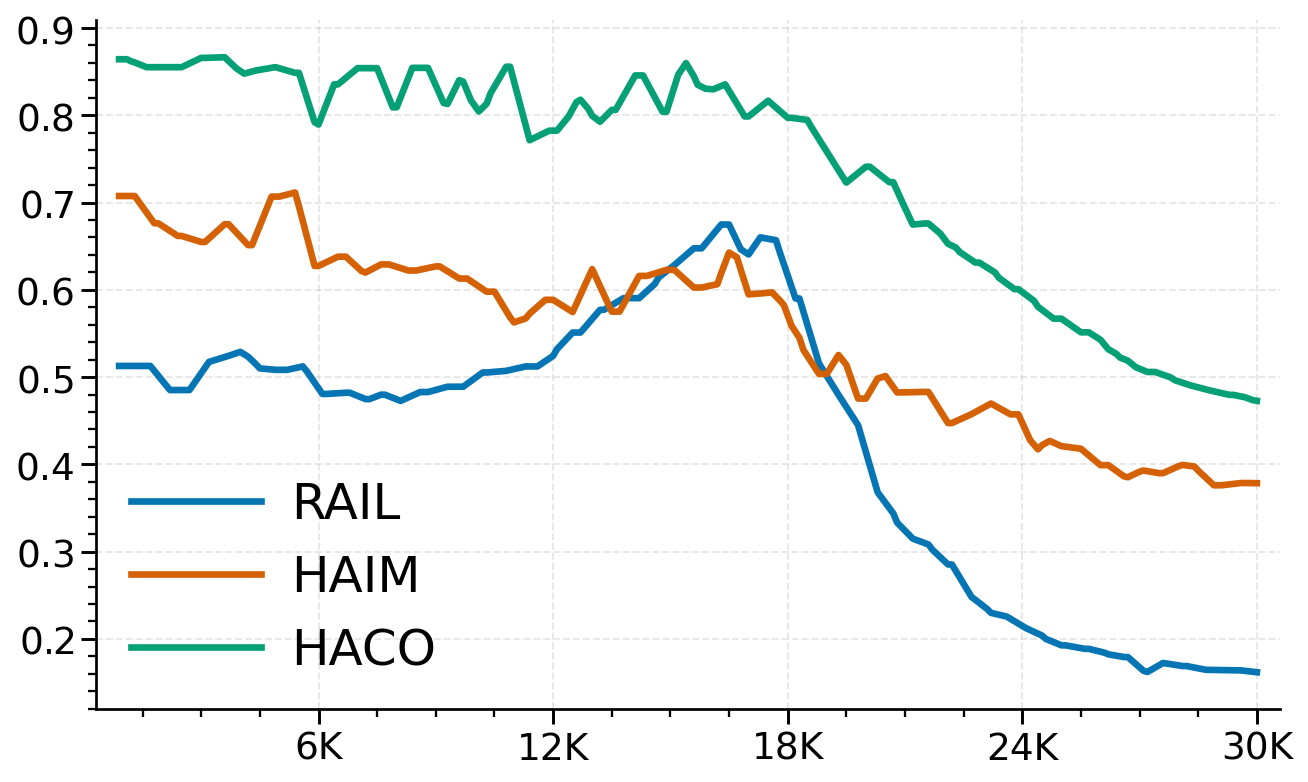}
  \caption{Takeover rate}
  \label{fig:tk_rate}
\end{subfigure}
\begin{subfigure}[t]{0.23\textwidth}
  \centering
  \includegraphics[width=\linewidth]{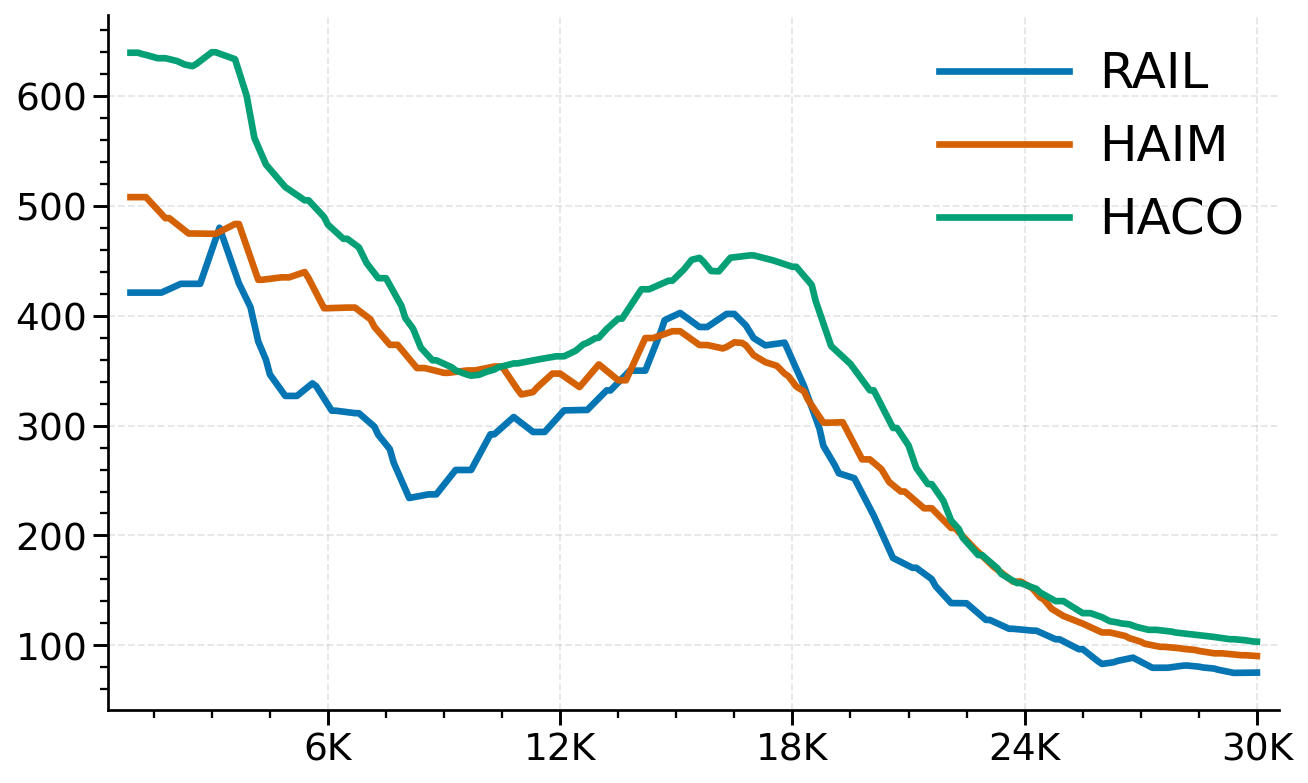}
  \caption{Takeover count}
  \label{fig:tk_count}
\end{subfigure}

\caption{Training Performance Comparison of RAIL, HAIM~\cite{huang2024human}, and HACO~\cite{li2022efficient} on MetaDrive}
\label{fig:rail_2x2}
\end{figure*}
\subsection{Risk-Prioritized Replay and Training Procedure}
RAIL concentrates learning on safety-critical experience by prioritizing risky or corrected transitions in replay. Each transition stores the core tuple $(s_t,u_t,r_t,s_{t+1})$ and attaches annotations $(a_t,\tilde a_t,\mathrm{IRS}_t,c_t,i^\star,k_t,\text{flags})$. A scalar priority combines critic TD error (learning progress), intrusion magnitude $\mathrm{IRS}_t$ (near misses), and a human-takeover flag (corrective supervision). Sampling is proportional to a powered priority, so batches remain diverse while emphasizing consequential events, and per-sample importance weights debias updates. Rollouts always run with risk-aware blending and contextual arbitration; even low-risk steps are retained with small priority to preserve nominal coverage. When $\mathrm{IRS}_t>\tau$, the log additionally records the active shield $k_t$, the instantaneous authority value $\alpha_t$ inside \text{flags}, and a short-horizon outcome tag (no-takeover vs.\ takeover). Training alternates brief interaction bursts to refresh bandit context with off-policy SAC updates from the prioritized buffer; in practice, human-takeover steps receive the highest priority, followed by high-$\mathrm{IRS}_t$ near misses, which accelerates safety alignment and reduces future interventions.

Fig.~\ref{fig:RAIL-arch} presents the RAIL system architecture. From state $s_t$, the policy $\pi_\theta$ proposes a nominal action $a_t$, which is assessed by three runtime risk cues—curvature uncertainty $r_{\mathrm{CURV}}$, time-to-collision $r_{\mathrm{TTC}}$, and observation shift $r_{\mathrm{OOD}}$—fused into the Intrusion Risk Score (IRS). When IRS exceeds a threshold, a contextual bandit selects a cue-specific shield $S_k$, and the final control $u_t$ is computed by blending $a_t$ with shield output $\tilde a_t$. Human overrides are possible at any time, and all transitions, annotations, and IRS values are logged into a risk-prioritized replay buffer to guide SAC learning.

\section{Experiments and Results}
We design our study around three guiding questions:
\begin{itemize}[leftmargin=2.5em]
    \item[\bf Q1:] \emph{How does \textbf{RAIL} compare to RL, safe RL, offline RL, IL, and HITL baselines in simulation across safety, performance, and traffic friendliness?}
    \item[\bf Q2:] \emph{How robust is \textbf{RAIL} to canonical cyber–physical intrusions relative to strong baselines?}
    \item[\bf Q3:] \emph{Which components of \textbf{RAIL} are most critical to its overall effectiveness?}
\end{itemize}

\subsection{Experimental Setup}

\paragraph{\bf Environment \& Scenarios}
We evaluate in \textsc{MetaDrive} with procedurally generated road graphs (straight segments, curves, T/X intersections, ramps, roundabouts, forks) populated by heterogeneous traffic from stochastic flows. Background vehicles follow IDM longitudinal dynamics and MOBIL lane changing; static cones/barriers and occasional slow movers induce long-tail interactions. Episodes end on goal, collision, off-road, or timeout (1{,}000 steps). For cross-simulator generalization, we also test in \textsc{CARLA} with a top-down semantic raster input and identical control heads.

\paragraph{\bf Threat Models}
\emph{Cyber attack: action–channel injection.} An adversary injects bounded deltas on steering and longitudinal acceleration after the policy but before the safety layer, modeled as $u_t=\mathrm{sat}(a_t+\eta_t)$. Piecewise-constant biases up to 50\% full-scale affect either channel for 5\,s every 30\,s (random start, sign, and channel), emulating ECU/CAN faults that cause unintended steering or throttle/brake bias and testing RAIL’s ability to contain lane departures and near misses. \emph{Physical attack: LiDAR spoofing.} A phantom obstacle is created by overwriting a contiguous azimuth sector in the 72-beam LiDAR with short ranges for a rigid object 4\,m ahead, with $\pm 0.5$\,m lateral jitter. Bursts last 5\,s and recur every 30\,s. The spoof feeds the environment's association, so TTC estimation perceives a plausible target. An attack is deemed successful if a collision or off-road occurs within 3\,s after a burst.

\paragraph{\bf Baselines}
We benchmark \textsc{RAIL} against representative methods spanning RL, safe RL, offline RL and IL, and HITL control. \emph{RL with reward shaping} includes SAC-RS \cite{haarnoja2018soft} and PPO-RS \cite{schulman2017proximal}, where “RS” denotes dense penalties on collisions, off-road, and excessive actuation added to the task reward. \emph{Safe RL with explicit cost constraints} comprises SAC-Lag \cite{ha2020learning} and PPO-Lag \cite{stooke2020responsive}, which optimize a Lagrangian relaxation with a learned multiplier on episodic cost, and CPO \cite{achiam2017constrained}, which enforces a trust-region update using a safety critic to satisfy a cost budget. \emph{Offline and imitation learning} baselines are CQL \cite{kumar2020conservative}, which regularizes Q-values toward conservatism on a fixed dataset, Behavior Cloning (BC) \cite{bain1995framework} that performs supervised learning on expert state–action pairs, and GAIL \cite{ho2016generative} that matches expert occupancy via adversarial training. \emph{Conventional HITL} methods include HG-Dagger \cite{kelly2019hg}, an interactive aggregation of human corrections into the training buffer, and IWR \cite{mandlekar2020human}, which upweights human takeover samples through intervention-weighted regression. We further compare to \emph{Human–AI copilot} HACO \cite{li2022efficient}, which blends online human guidance with policy optimization, and \emph{Human as AI Mentor} HAIM-DRL \cite{huang2024human}, which treats the expert as an online mentor providing corrective signals during exploration. All baselines use the same observation encoder, action bounds, control rate, seeds, and training budget as \textsc{RAIL}, with algorithm-specific defaults preserved where required by the original papers.

Fig.~\ref{fig:rail_2x2} shows training curves on MetaDrive comparing RAIL with HAIM and HACO. RAIL consistently lowers disturbance and operator takeovers faster than baselines, reaching the lowest steady-state levels across all metrics.

\paragraph{\bf Metrics}
We evaluate performance and robustness using formally defined, episode-aggregated metrics:
\begin{itemize}
    \item \textbf{Test Success Rate (TSR):} fraction of episodes reaching the goal without collisions or off-road events.
    \item \textbf{Test Return (TR):} expected cumulative reward, reflecting task efficiency.
    \item \textbf{Test Safety Violations (TSV):} number of collisions or lane departures per episode.
    \item \textbf{Disturbance Rate (DR):} fraction of control ticks where acceleration exceeds a comfort bound.

    \item \textbf{Training Safety Violations (TSV$_{\text{train}}$):} cumulative number of safety events during training.
    \item \textbf{Training Data Usage (TDU):} total number of interaction steps during training.
    \item \textbf{Disengagement Rate under Attack (DRA):} fraction of attacked episodes with human takeover.
    \item \textbf{Attack Success Rate (ASR):} fraction of attacked episodes ending in collision or off-road.
\end{itemize}


\begin{table*}[t!]
\centering
\caption{Performance comparison of RL, safe RL, offline RL, imitation learning (IL), and human-in-the-loop (HITL) methods on MetaDrive. Values are mean $\pm$ standard deviation over five seeds; best results are in \textbf{bold}.}
\label{tab:metadrive_results}
\begin{adjustbox}{width=1.0\textwidth}

\begin{tabular}{llcccccc}
\toprule
Category & Method 
& \makecell{Total Training\\Safety Violation} 
& \makecell{Training\\Data Usage} 
& \makecell{Test\\Return} 
& \makecell{Test Safety\\Violation} 
& \makecell{Disturbance\\Rate} 
& \makecell{Test Success\\Rate} \\
\midrule
\emph{Expert} 
    & Human & -- & -- & \meanstd{388.16}{45.00} & \meanstd{0.03}{0.00} & 0 & 1.00 \\
\midrule
\multirow{2}{*}{RL} 
    & SAC-RS \cite{haarnoja2018soft} & \meanstd{2.78K}{0.97K} & 1M & \textbf{\meanstd{384.56}{37.5}} & \meanstd{0.87}{1.47} & \meanstd{0.015}{0.005} & \meanstd{0.83}{0.32} \\
    & PPO-RS \cite{schulman2017proximal} & \meanstd{27.51K}{3.86K} & 1M & \meanstd{305.41}{14.23} & \meanstd{4.12}{1.24} & \meanstd{0.021}{0.014} & \meanstd{0.67}{0.12} \\
\midrule
\multirow{3}{*}{Safe RL} 
    & SAC-Lag \cite{ha2020learning} & \meanstd{1.98K}{0.75K} & 1M & \meanstd{352.46}{108.78} & \meanstd{0.78}{0.58} & \meanstd{0.019}{0.007} & \meanstd{0.71}{0.79} \\
    & PPO-Lag \cite{stooke2020responsive} & \meanstd{15.46K}{5.13K} & 1M & \meanstd{298.98}{50.99} & \meanstd{3.28}{0.38} & \meanstd{0.025}{0.016} & \meanstd{0.52}{0.27} \\
    & CPO \cite{achiam2017constrained} & \meanstd{4.36K}{2.22K} & 1M & \meanstd{194.06}{108.86} & \meanstd{1.71}{1.02} & -- & \meanstd{0.21}{0.29} \\
\midrule
Offline RL 
    & CQL \cite{kumar2020conservative} & -- & 49K & \meanstd{116.45}{34.94} & \meanstd{3.68}{7.61} & \meanstd{0.007}{0.006} & \meanstd{0.13}{0.09} \\
\midrule
\multirow{2}{*}{IL} 
    & BC \cite{bain1995framework} & -- & 49K & \meanstd{36.13}{10.06} & \meanstd{1.05}{0.54} & \meanstd{0.012}{0.017} & \meanstd{0.01}{0.02} \\
    & GAIL \cite{ho2016generative} & \meanstd{3.68K}{3.17K} & 49K & \meanstd{108.36}{16.08} & \meanstd{4.18}{1.25} & \meanstd{0.001}{0.0009} & \meanstd{0.03}{0.01} \\
\midrule
\multirow{2}{*}{Conventional HITL}
    & HG-Dagger \cite{kelly2019hg} & 35.58 & 50K & 106.21 & 2.63 & 0.108 & 0.04 \\
    & IWR \cite{mandlekar2020human} & 69.74 & 50K & 298.87 & 3.61 & 0.122 & 0.61 \\
\midrule
Human-AI Copilot 
    & HACO \cite{li2022efficient} & \meanstd{30.05}{10.89} & 30K & \meanstd{350.01}{9.72} & \meanstd{0.78}{0.85} & \meanstd{0.038}{0.0083} & \meanstd{0.83}{0.07} \\
\midrule
Human as AI Mentor 
    & HAIM-DRL \cite{huang2024human} & \meanstd{29.84}{10.25} & 30K & \meanstd{354.34}{11.08} & \meanstd{0.76}{0.28} & \textbf{\meanstd{0.0023}{0.00072}} & \textbf{\meanstd{0.85}{0.03}} \\
\midrule
\rowcolor{lightgray}
\textbf{Risk-Aware HITL} & \textbf{RAIL (Ours)} & \textbf{\meanstd{29.07}{11.36}} & \textbf{30K} & \text{\meanstd{360.65}{23.06}} & \textbf{\meanstd{0.75}{0.03}} & \meanstd{0.0027}{0.00052} & \textbf{\meanstd{0.85}{0.07}} \\
\bottomrule
\end{tabular}
\end{adjustbox}
\end{table*}

\paragraph{\bf Implementation details}
All methods use SAC with shared hyperparameters (selected through careful tuning using grid search): actor, critic, and entropy learning rates of $1\times10^{-4}$, batch size 1024, horizon 1{,}000, learning starts at 100 steps, control at 10 Hz, action clipping, and first-order rate limiting. Observations include ego kinematics, 72-beam LiDAR with 60 m range, side detectors, lane features, and diagnostics used by \textsc{RAIL}. Actions 
$a_t=[a_t^{\text{steer}},a_t^{\text{acc}}]$
lie in $[-1,1]^2$ and map to $\pm0.35$ rad steering and $[-4.0,2.5]$ m s$^{-2}$. The Intrusion Risk Score uses weights $(0.3,0.4,0.3)$ with threshold $\tau=0.3$. When the score exceeds $\tau$, a contextual selector chooses a shield and the executor blends with authority $\alpha_t$ in $[0,1]$. Each method is trained with five random seeds up to a fixed budget of environment interactions or until a validation return plateau, and we keep the best checkpoint on the validation split.

\subsection{Performance Comparison with Baselines}

Table~\ref{tab:metadrive_results} compares reward-shaped RL, safe RL, offline RL, imitation learning (IL), and HITL methods. Reward-shaped RL attains a high return but weaker safety: SAC-RS reaches 384.56 return with 0.87 test safety violations and 0.015 disturbance, while PPO-RS drops to 305.41 return with 4.12 violations and 0.021 disturbance. Safe RL reduces violations relative to PPO-RS but also lowers return and success rate; for instance, SAC-Lag achieves 352.46 return with 0.78 violations and 0.019 disturbance, whereas PPO-Lag yields 298.98 return with 3.28 violations and 0.025 disturbance. Offline and imitation learning baselines perform poorly in this setting: CQL obtains 116.45 return with 3.68 violations and 0.007 disturbance, BC yields 36.13 return with 1.05 violations and 0.012 disturbance, and GAIL reaches 108.36 return with 4.18 violations and 0.001 disturbance. HITL methods close the gap: HACO achieves 350.01 return with 0.78 violations and 0.038 disturbance, while HAIM-DRL attains 354.34 return with 0.76 violations and 0.0023 disturbance.

RAIL delivers a favorable balance of safety and efficiency with only 30K interaction steps. It achieves 360.65 return and matches the top success rate at 0.85, while reducing test safety violations to 0.75 and keeping disturbance at 0.0027. Training safety violations are also low at 29.07, indicating safer data collection. Compared to HAIM-DRL, RAIL improves return, maintains comparable disturbance and success, and substantially outperforms reward-shaped PPO and SAC on safety. These results show that RAIL preserves HITL sample efficiency, improves safety without sacrificing task performance, and narrows the gap to expert-level stability. A similar trend holds in \textsc{CARLA}, where RAIL attains 1609.70 return and 0.41 success rate with only 8K samples (Table~\ref{tab:carla_gen}).

\begin{table}[t]
\centering
\caption{\centering CARLA simulator generalization. }
\label{tab:carla_gen}
\small
\renewcommand{\arraystretch}{1.05}
\setlength{\tabcolsep}{4pt}
\begin{adjustbox}{width=\columnwidth}
\begin{tabular}{lcccc}
\toprule
\textbf{Method} & \textbf{Training Data} & \textbf{TSV$\downarrow$} & \textbf{TR $\uparrow$} & \textbf{TSR$\uparrow$} \\
\midrule
PPO \cite{schulman2017proximal}        & 500K  & 80.84 & 1591.00 & 0.35 \\
HACO \cite{li2022efficient}       & 8K   & 12.14 & 1578.43 & 0.35 \\
HAIM \cite{huang2024human}  & 8K    & \textbf{11.25} & 1590.85 & 0.38 \\
\midrule
\textbf{RAIL (Ours)} & \textbf{8K} & 12.38 & \textbf{1609.70} & \textbf{0.41} \\
\bottomrule
\end{tabular}
\end{adjustbox}

\vspace{1mm}
(Note: TSV$\downarrow$=test safety violations; TR=test return; TSR$\uparrow$=test success rate.)
\end{table}

\subsection{Robustness to Intrusions}

\begin{table}[t]
\centering
\caption{\centering Robustness on MetaDrive under cyber and physical \emph{intrusion} attacks. }
\renewcommand{\arraystretch}{1.08}
\small
\setlength{\tabcolsep}{5pt}

\begin{adjustbox}{width=0.48\textwidth} 
\begin{tabular}{lcc}
\toprule
\textbf{Method} & \textbf{CAN Injection} & \textbf{LiDAR Spoof} \\
\midrule
\textit{} & \emph{TSR$\uparrow$ / DRA$\downarrow$ / ASR$\downarrow$} & \emph{TSR$\uparrow$ / DRA$\downarrow$ / ASR$\downarrow$} \\
\midrule
PPO \cite{schulman2017proximal}        & 0.23 / 0.85 / 0.62 & 0.49 / 0.21 / 0.23 \\
HACO \cite{li2022efficient}        & 0.58 / 0.90 / 0.75 & 0.74 / 0.24 / 0.29 \\
HAIM \cite{huang2024human}    & 0.62 / 0.74 / 0.65 & 0.77 / 0.17 / 0.20 \\
\midrule
\textbf{RAIL (Ours)} & \textbf{0.68 / 0.37 / 0.34} & \textbf{0.80 / 0.03 / 0.11} \\
\bottomrule
\end{tabular}
\end{adjustbox}
\label{tab:robustness-fixed}

\vspace{1mm}
(Note: Each cell reports \textbf{TSR} (Test Success Rate), \textbf{DRA} (Disengagement Rate under Attack), and \textbf{ASR} (Attack Success Rate).)
\end{table}

Table~\ref{tab:robustness-fixed} shows that \textbf{RAIL} is the most robust under both intrusion settings. Under CAN injection, RAIL attains the highest test success rate (TSR \(0.68\)) versus PPO \(0.23\), HACO \(0.58\), and HAIM \(0.62\). It also cuts disengagements (DRA \(0.37\)) far below HACO \(0.90\) and HAIM \(0.74\), while almost halving attack success (ASR \(0.34\) vs.\ HAIM \(0.65\) and PPO \(0.62\)). Under LiDAR spoofing, RAIL again leads with SR \(0.80\), achieves near-zero DRA \(0.03\) compared to HAIM \(0.17\) and HACO \(0.24\), and reduces ASR to \(0.11\) from HAIM \(0.20\) and HACO \(0.29\). These outcomes indicate that RAIL’s cue-driven shielding contains actuation corruption and phantom obstacles without collapsing into conservative behavior or relying on frequent human takeover. The simultaneous gains in SR and reductions in DRA and ASR suggest effective containment rather than avoidance, lowering operator burden while preserving task performance in contested conditions.

\subsection{Ablation Study of RAIL Components}

\begin{table}[t]
\centering
\caption{RAIL ablations on MetaDrive.}
\small
\renewcommand{\arraystretch}{1.05}
\setlength{\tabcolsep}{1.8pt}
\begin{tabular}{lccccc}
\toprule
\textbf{Variant} & \textbf{SR$\uparrow$} & \textbf{SV$\downarrow$} & \textbf{DR$\downarrow$} & \textbf{AIRS$\downarrow$} & \textbf{AS$\uparrow$} \\
\midrule
w/o curvature cue           & 0.78 & 0.80 & 0.0140 & 0.17 & 22.7 \\
w/o TTC cue                 & 0.67 & 1.02 & 0.0220 & \textbf{0.12} & 22.5 \\
w/o OOD cue                 & 0.83 & 0.77 & 0.0049 & 0.15 & \textbf{23.3} \\
w/o bandit (fixed)          & 0.79 & 1.10 & 0.0180 & 0.16 & 19.7 \\
w/o weights (uniform)       & 0.81 & 0.79 & 0.0056 & 0.16 & 20.4 \\
w/o shield penalty          & 0.74 & \textbf{0.72} & 0.0044 & \textbf{0.12} & 16.6 \\
\midrule
\textbf{RAIL (full)}        & \textbf{0.85} & 0.75 & \textbf{0.0027} & 0.15 & 22.8 \\
\bottomrule
\end{tabular}
\label{tab:RAIL-ablation-nominal}
\vspace{1mm}
(Note: SR$\uparrow$ = success rate; SV$\downarrow$ = safety violations; DR$\downarrow$ = disturbance rate; 
AIRS$\downarrow$ = average Intrusion Risk Score per episode; AS$\uparrow$ = average speed (km/h))
\end{table}

Table~\ref{tab:RAIL-ablation-nominal} indicates that \textbf{TTC} is the most critical cue: removing it degrades SR to \(0.67\) and sharply increases both SV (\(1.02\)) and DR (\(0.022\)), confirming its role in imminent-collision mitigation. The \textbf{curvature} cue is next in importance (SR \(0.78\), SV \(0.80\), DR \(0.014\)), reflecting its contribution to actuation integrity. Dropping the \textbf{OOD} cue yields near-normal SR (\(0.83\)) but higher DR (\(0.0049\)) and slightly worse SV, suggesting reduced caution at higher speeds (AS \(23.3\)). Adaptive \textbf{bandit} arbitration matters: fixed shields raise SV (\(1.10\)) and DR (\(0.018\)) while lowering AS. Uniform IRS weights also hurt DR and AS. Finally, removing the \textbf{shield penalty} induces over-use of shielding—SR falls to \(0.74\) and AS to \(16.6\), showing the penalty is needed to prevent overly conservative behavior while maintaining low DR.

\begin{figure}[t]
\centering

\begin{subfigure}[t]{0.32\linewidth}
  \centering
  \includegraphics[width=\linewidth]{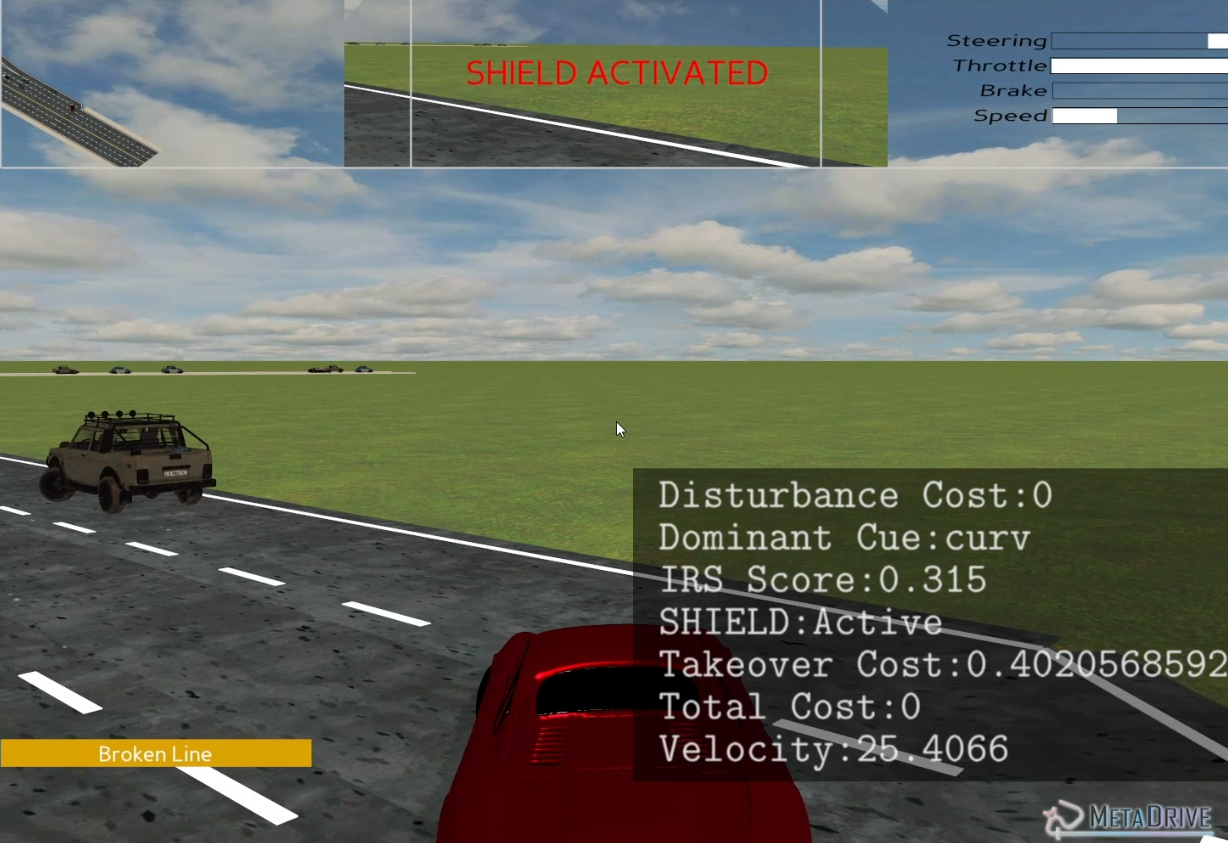}\\[3pt]
  \includegraphics[width=\linewidth]{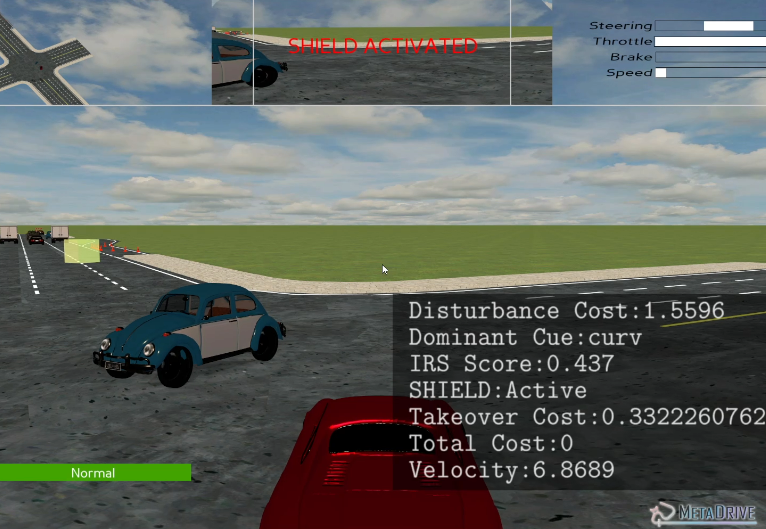} 
  \caption{Curvature}
\end{subfigure}\hfill
\begin{subfigure}[t]{0.32\linewidth}
  \centering
  \includegraphics[width=\linewidth]{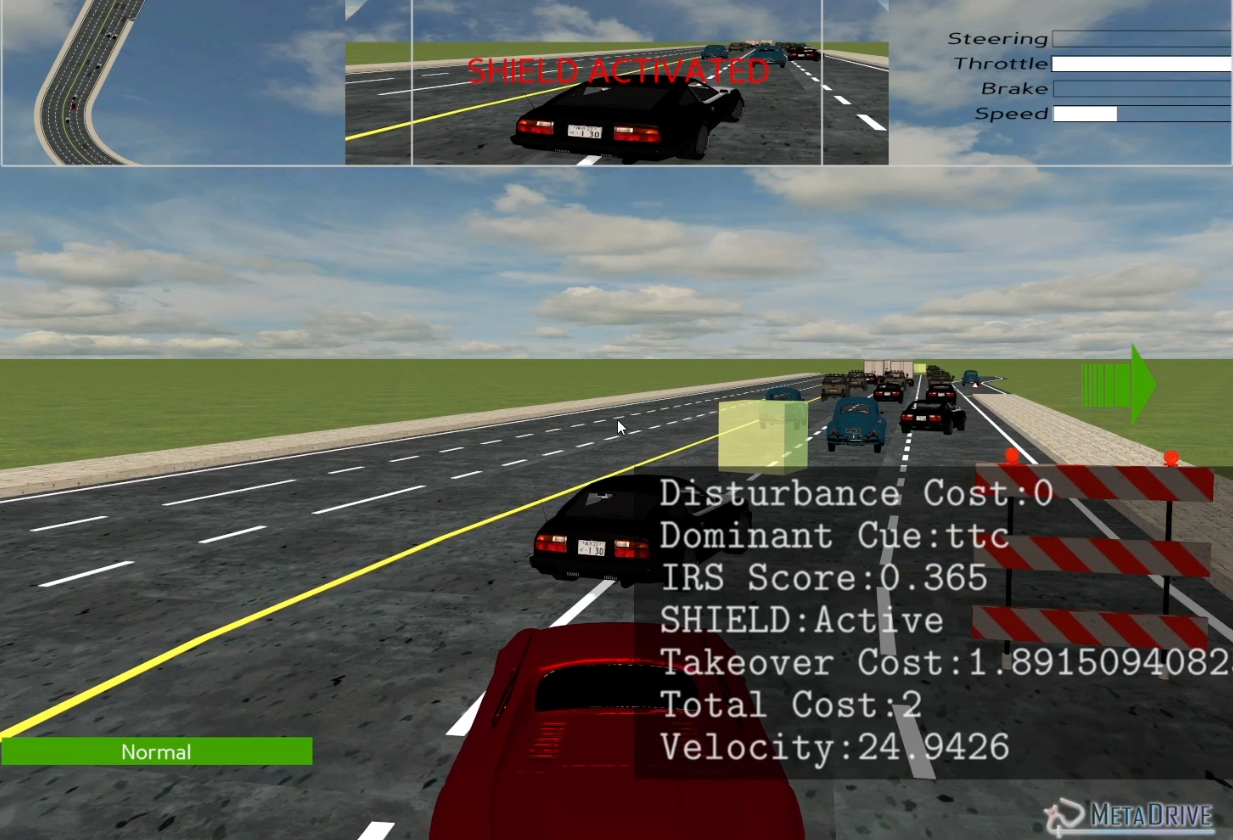}\\[3pt]
  \includegraphics[width=\linewidth]{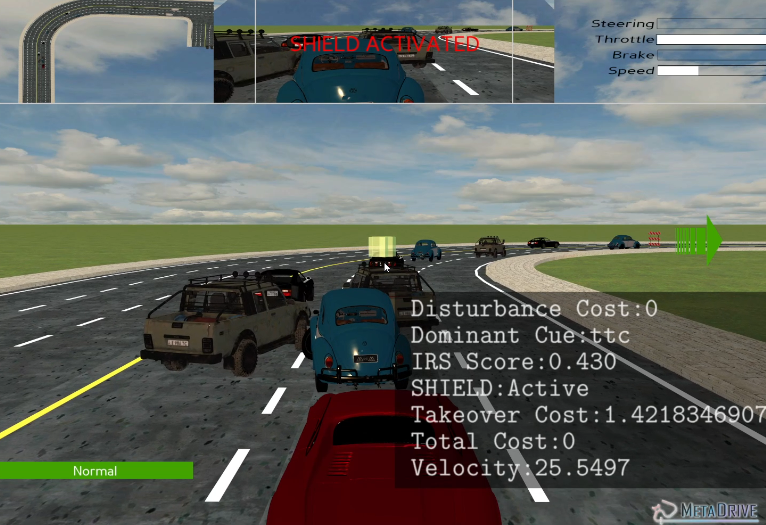} 
  \caption{TTC}
\end{subfigure}\hfill
\begin{subfigure}[t]{0.32\linewidth}
  \centering
  \includegraphics[width=\linewidth]{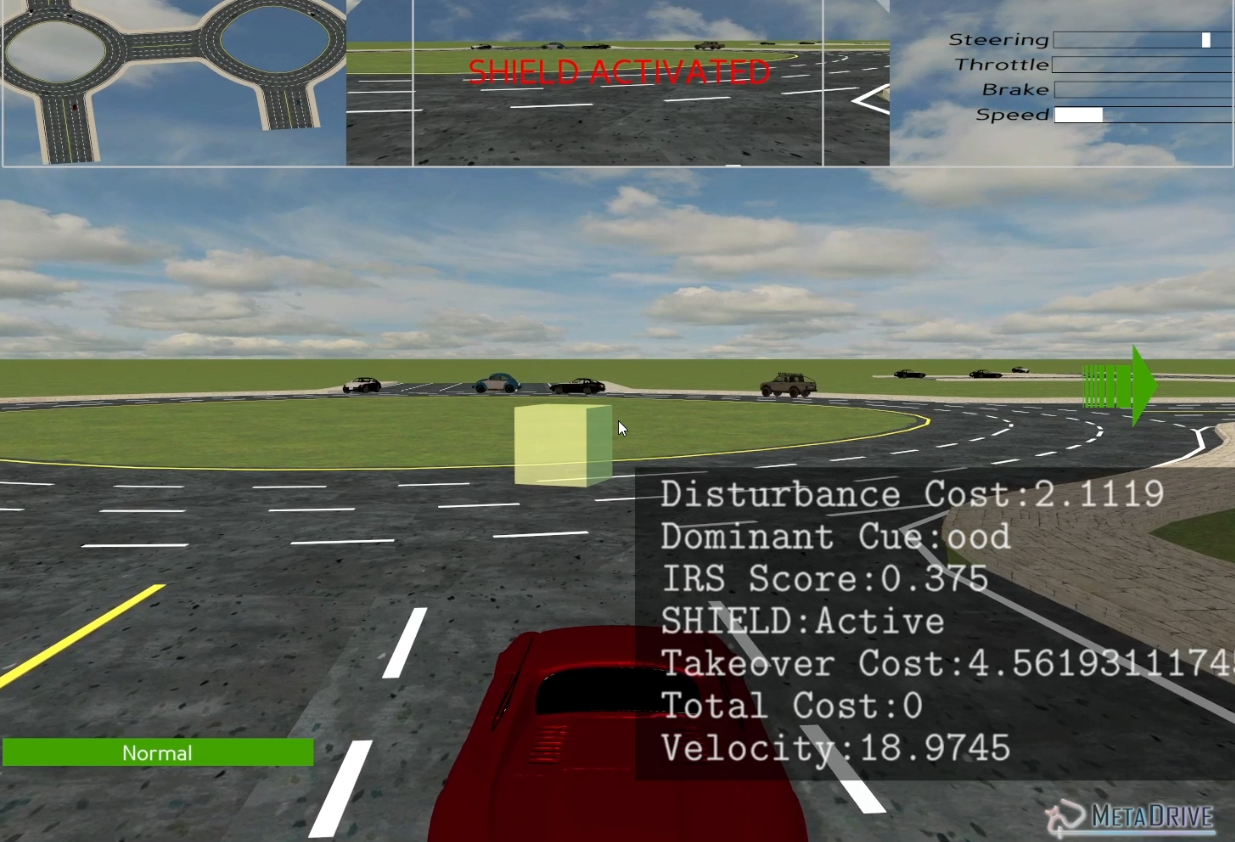}\\[3pt]
  \includegraphics[width=\linewidth]{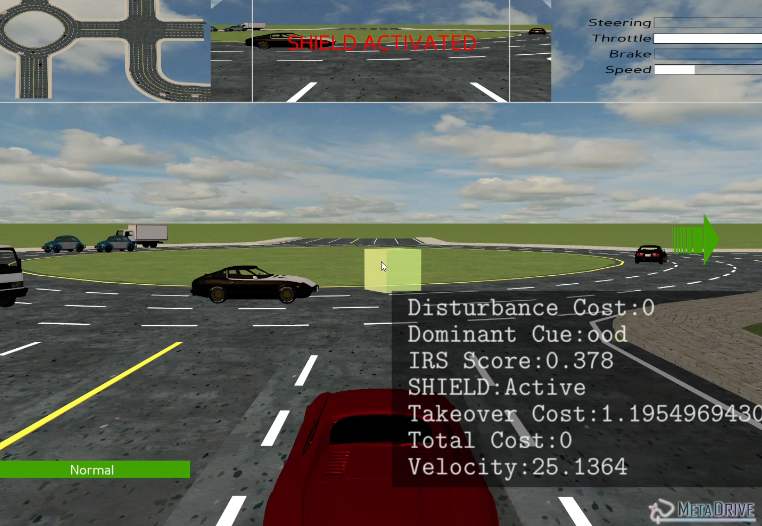} 
  \caption{OOD}
\end{subfigure}

\caption{Qualitative Cue-Driven IRS Responses}
\label{fig:rail_qual_2x3}
\end{figure}

Fig.~\ref{fig:rail_qual_2x3} illustrates qualitative examples where IRS spikes trigger dominant cue selection and shielded controls. For each cue (curvature, TTC, OOD), RAIL responds with targeted interventions that avert incidents while maintaining driving progress.



\section{Conclusions \& Future Work}
\label{sec:conclusions}

We presented \textsc{RAIL}, a risk-aware human-in-the-loop framework that fuses heterogeneous runtime cues into an Intrusion Risk Score, blends nominal and safeguarded controls via interpretable shields, arbitrates with a contextual bandit, and learns with dual rewards plus risk-prioritized replay. On MetaDrive, RAIL improves safety and robustness over RL, safe RL, imitation, offline RL, and prior HITL baselines, lowering disturbance and disengagement while maintaining competitive success and return. Under cyber and physical intrusions, RAIL lowers attack success and the need for operator handovers, demonstrating that intrusion-aware shaping plus online shield adaptation translates concern into graded control actions, preserving stability and mission progress rather than triggering an all-or-nothing fail-safe.

For \textbf{future work}, we will broaden cue coverage (e.g., richer perception consistency checks), study adaptive thresholds and shield parameterization with formal safety envelopes, and scale to multi-agent traffic with cooperative risk sharing. We also plan to transfer to hardware-in-the-loop tests, expand human factors evaluation of transparency and workload, and integrate stronger attack generation to stress-test generalization.



\bibliographystyle{IEEEtran}
\bibliography{main}

@inproceedings{haarnoja2018soft,
  title={Soft actor-critic: Off-policy maximum entropy deep reinforcement learning with a stochastic actor},
  author={Haarnoja, Tuomas and Zhou, Aurick and Abbeel, Pieter and Levine, Sergey},
  booktitle={International conference on machine learning},
  pages={1861--1870},
  year={2018},
  organization={Pmlr}
}

@article{schulman2017proximal,
  title={Proximal policy optimization algorithms},
  author={Schulman, John and Wolski, Filip and Dhariwal, Prafulla and Radford, Alec and Klimov, Oleg},
  journal={arXiv preprint arXiv:1707.06347},
  year={2017}
}

@article{ha2020learning,
  title={Learning to walk in the real world with minimal human effort},
  author={Ha, Sehoon and Xu, Peng and Tan, Zhenyu and Levine, Sergey and Tan, Jie},
  journal={arXiv preprint arXiv:2002.08550},
  year={2020}
}

@inproceedings{stooke2020responsive,
  title={Responsive safety in reinforcement learning by pid lagrangian methods},
  author={Stooke, Adam and Achiam, Joshua and Abbeel, Pieter},
  booktitle={International Conference on Machine Learning},
  pages={9133--9143},
  year={2020},
  organization={PMLR}
}

@inproceedings{achiam2017constrained,
  title={Constrained policy optimization},
  author={Achiam, Joshua and Held, David and Tamar, Aviv and Abbeel, Pieter},
  booktitle={International conference on machine learning},
  pages={22--31},
  year={2017},
  organization={PMLR}
}

@article{kumar2020conservative,
  title={Conservative q-learning for offline reinforcement learning},
  author={Kumar, Aviral and Zhou, Aurick and Tucker, George and Levine, Sergey},
  journal={Advances in neural information processing systems},
  volume={33},
  pages={1179--1191},
  year={2020}
}

@inproceedings{bain1995framework,
  title={A Framework for Behavioural Cloning.},
  author={Bain, Michael and Sammut, Claude},
  booktitle={Machine intelligence 15},
  pages={103--129},
  year={1995}
}

@article{ho2016generative,
  title={Generative adversarial imitation learning},
  author={Ho, Jonathan and Ermon, Stefano},
  journal={Advances in neural information processing systems},
  volume={29},
  year={2016}
}

@inproceedings{kelly2019hg,
  title={Hg-dagger: Interactive imitation learning with human experts},
  author={Kelly, Michael and Sidrane, Chelsea and Driggs-Campbell, Katherine and Kochenderfer, Mykel J},
  booktitle={2019 International Conference on Robotics and Automation (ICRA)},
  pages={8077--8083},
  year={2019},
  organization={IEEE}
}

@article{mandlekar2020human,
  title={Human-in-the-loop imitation learning using remote teleoperation},
  author={Mandlekar, Ajay and Xu, Danfei and Mart{\'\i}n-Mart{\'\i}n, Roberto and Zhu, Yuke and Fei-Fei, Li and Savarese, Silvio},
  journal={arXiv preprint arXiv:2012.06733},
  year={2020}
}

@article{li2022efficient,
  title={Efficient learning of safe driving policy via human-ai copilot optimization},
  author={Li, Quanyi and Peng, Zhenghao and Zhou, Bolei},
  journal={arXiv preprint arXiv:2202.10341},
  year={2022}
}

@article{huang2024human,
  title={Human as AI mentor: Enhanced human-in-the-loop reinforcement learning for safe and efficient autonomous driving},
  author={Huang, Zilin and Sheng, Zihao and Ma, Chengyuan and Chen, Sikai},
  journal={Communications in Transportation Research},
  volume={4},
  pages={100127},
  year={2024},
  publisher={Elsevier}
}

@inproceedings{alshiekh2018safe,
  title={Safe reinforcement learning via shielding},
  author={Alshiekh, Mohammed and Bloem, Roderick and Ehlers, R{\"u}diger and K{\"o}nighofer, Bettina and Niekum, Scott and Topcu, Ufuk},
  booktitle={Proceedings of the AAAI conference on artificial intelligence},
  volume={32},
  number={1},
  year={2018}
}

@inproceedings{tamar2015optimizing,
  title={Optimizing the CVaR via sampling},
  author={Tamar, Aviv and Glassner, Yonatan and Mannor, Shie},
  booktitle={Proceedings of the AAAI Conference on Artificial Intelligence},
  volume={29},
  number={1},
  year={2015}
}

@article{greenberg2022efficient,
  title={Efficient risk-averse reinforcement learning},
  author={Greenberg, Ido and Chow, Yinlam and Ghavamzadeh, Mohammad and Mannor, Shie},
  journal={Advances in Neural Information Processing Systems},
  volume={35},
  pages={32639--32652},
  year={2022}
}

@article{queeney2023risk,
  title={Risk-averse model uncertainty for distributionally robust safe reinforcement learning},
  author={Queeney, James and Benosman, Mouhacine},
  journal={Advances in Neural Information Processing Systems},
  volume={36},
  pages={1659--1680},
  year={2023}
}

@inproceedings{yang2024risk,
  title={Risk-aware constrained reinforcement learning with non-stationary policies},
  author={Yang, Zhaoxing and Jin, Haiming and Tang, Yao and Fan, Guiyun},
  booktitle={Proceedings of the 23rd International Conference on Autonomous Agents and Multiagent Systems},
  pages={2029--2037},
  year={2024}
}

@inproceedings{cho2016fingerprinting,
  title={Fingerprinting electronic control units for vehicle intrusion detection},
  author={Cho, Kyong-Tak and Shin, Kang G},
  booktitle={25th USENIX security symposium (USENIX Security 16)},
  pages={911--927},
  year={2016}
}

@article{kastner2023distributional,
  title={Distributional model equivalence for risk-sensitive reinforcement learning},
  author={Kastner, Tyler and Erdogdu, Murat A and Farahmand, Amir-massoud},
  journal={Advances in Neural Information Processing Systems},
  volume={36},
  pages={56531--56552},
  year={2023}
}

@article{tamar2015policy,
  title={Policy gradient for coherent risk measures},
  author={Tamar, Aviv and Chow, Yinlam and Ghavamzadeh, Mohammad and Mannor, Shie},
  journal={Advances in neural information processing systems},
  volume={28},
  year={2015}
}

@article{sultana2024detects,
  title={LA-DETECTS: Local and Adaptive Data-Centric Misbehavior Detection Framework for Vehicular Technology Security},
  author={Sultana, Rukhsar and Grover, Jyoti and Tripathi, Meenakshi and Sharma, Prinkle},
  journal={IEEE Open Journal of Vehicular Technology},
  year={2024},
  publisher={IEEE}
}

@inproceedings{ren2020adversarial,
  title={Adversarial example attacks in the physical world},
  author={Ren, Huali and Huang, Teng},
  booktitle={International Conference on Machine Learning for Cyber Security},
  pages={572--582},
  year={2020},
  organization={Springer}
}

@article{aloraini2024adversarial,
  title={Adversarial attacks on intrusion detection systems in in-vehicle networks of connected and autonomous vehicles},
  author={Aloraini, Fatimah and Javed, Amir and Rana, Omer},
  journal={Sensors},
  volume={24},
  number={12},
  pages={3848},
  year={2024},
  publisher={MDPI}
}

@article{hamad2024react,
  title={REACT: Autonomous intrusion response system for intelligent vehicles},
  author={Hamad, Mohammad and Finkenzeller, Andreas and K{\"u}hr, Michael and Roberts, Andrew and Maennel, Olaf and Prevelakis, Vassilis and Steinhorst, Sebastian},
  journal={Computers \& Security},
  volume={145},
  pages={104008},
  year={2024},
  publisher={Elsevier}
}

@inproceedings{abdo2024avmon,
  title={Avmon: securing autonomous vehicles by learning control invariants and residual prediction},
  author={Abdo, Ahmed and Malek, Sakib Md Bin and Zhao, Xuanpeng and Abu-Ghazaleh, Nael},
  booktitle={Symposium on Vehicle Security and Privacy (VehicleSec)},
  year={2024}
}

@article{nagarajan2023machine,
  title={Machine Learning based intrusion detection systems for connected autonomous vehicles: A survey},
  author={Nagarajan, Jay and Mansourian, Pegah and Shahid, Muhammad Anwar and Jaekel, Arunita and Saini, Ikjot and Zhang, Ning and Kneppers, Marc},
  journal={Peer-to-Peer Networking and Applications},
  volume={16},
  number={5},
  pages={2153--2185},
  year={2023},
  publisher={Springer}
}

@article{parekh2022review,
  title={A review on autonomous vehicles: Progress, methods and challenges},
  author={Parekh, Darsh and Poddar, Nishi and Rajpurkar, Aakash and Chahal, Manisha and Kumar, Neeraj and Joshi, Gyanendra Prasad and Cho, Woong},
  journal={Electronics},
  volume={11},
  number={14},
  pages={2162},
  year={2022},
  publisher={MDPI}
}

@article{saunders2017trial,
  title={Trial without error: Towards safe reinforcement learning via human intervention},
  author={Saunders, William and Sastry, Girish and Stuhlmueller, Andreas and Evans, Owain},
  journal={arXiv preprint arXiv:1707.05173},
  year={2017}
}

@article{pendleton2016survey,
  title={A survey on systems security metrics},
  author={Pendleton, Marcus and Garcia-Lebron, Richard and Cho, Jin-Hee and Xu, Shouhuai},
  journal={ACM Computing Surveys (CSUR)},
  volume={49},
  number={4},
  pages={1--35},
  year={2016},
  publisher={ACM New York, NY, USA}
}

@article{johansen2014foundations,
  title={Foundations and choice of risk metrics},
  author={Johansen, Inger Lise and Rausand, Marvin},
  journal={Safety science},
  volume={62},
  pages={386--399},
  year={2014},
  publisher={Elsevier}
}

@inproceedings{choudhary2018intrusion,
  title={Intrusion detection systems for networked unmanned aerial vehicles: A survey},
  author={Choudhary, Gaurav and Sharma, Vishal and You, Ilsun and Yim, Kangbin and Chen, Ray and Cho, Jin-Hee},
  booktitle={2018 14th International Wireless Communications \& Mobile Computing Conference (IWCMC)},
  pages={560--565},
  year={2018},
  organization={IEEE}
}

\end{document}